# Cursive Overlapped Character Segmentation: An Enhanced Approach

Amjad Rehman

College of Business Administration Al Yamamah University Riyadh Saudi Arabia

**Abstract**

Segmentation of highly slanted and horizontally overlapped characters is a challenging research area that is still fresh. Several techniques are reported in the state of art, but produce low accuracy for the highly slanted characters segmentation and cause overall low handwriting recognition precision. Accordingly, this paper presents a simple yet effective approach for character segmentation of such difficult slanted cursive words without using any slant correction technique. Rather a new concept of core-zone is introduced for segmenting such difficult slanted handwritten words. However, due to the inherent nature of cursive words, few characters are over-segmented and therefore, a threshold is selected heuristically to overcome this problem. For fair comparison, difficult words are extracted from the IAM benchmark database. Experiments thus performed exhibit promising result and high speed.

## 1    Introduction and background

Segmentation of handwritten words is a long-standing problem at the heart of systems for conversion of handwritten information to electronic form [1-5]. This problem becomes crucial for highly slanted and horizontally overlapped cursive words. The literature is replete with high accuracy recognition for separated handwritten numerals and characters [6-12]. However, segmentation and recognition of characters extracted from cursive, touching and horizontally overlapped do not have the same measure of success [13-18]. Accurate segmentation is the difficult phase for correct and speedy recognition, particularly in the context of segmentation-based, word recognition [19-22]. Still, there are several reasons why we may find an advantage in more accurate and reliable word image segmentation. Firstly, for the reason of efficiency, we would like to keep the number of segmentations as less as possible. Secondly, each possible segmentation point in the image requires some resolution of target character, which could extend the time of character image recognition significantly [23-27].

The segmentation is the backbone of the recognition process and is still an active research area. Hence, higher character segmentation accuracy is directly proportional to handwriting recognition accuracy [28-32]. Researchers have acknowledged the important role that segmentation plays in the handwriting recognition process [33-37]. That is why more innovative, accurate and fast techniques need to be employed and compared to the work of other researchers using benchmark databases [38,39]. However, the words that are highly slanted and overlapped horizontally cannot be correctly segmented despite slant correction techniques [40,41].

This paper presents a simple but effective approach for segmenting such difficult cursive words extracted from benchmark dataset. In the proposed approach, rather performing complex slant correction preprocessing techniques, word core-zone height is detected. Segmentation performed in the core zone yields better results with high speed. Once character boundaries are found, difficult slanted and horizontally overlapped characters can be detected by following connected pixels.

Rest of the paper is organized in three main sections. Sections 2 presents the proposed segmentation approach for difficult words. In section 3 experimental results are exhibited and discussed. Finally, conclusion is drawn in section 4.

## 2   Proposed segmentation approach

In this section, preprocessing steps and segmentation algorithm is presented. An overview of the proposed segmentation algorithm is shown in figure 1

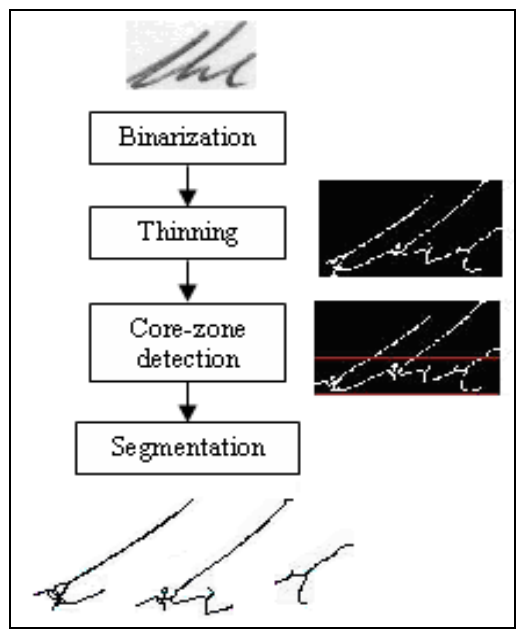

Figure 1. Overview of the segmentation algorithm

### 2.1   Binarization

The first step removes noise using the Otsu algorithm to change the original image into binary form [41]. The output binary image has values 1 (for white pixels) and values 0 (for black pixels).

### 2.2   Thinning

Following binarization, the image is converted to skeleton image to allow users verity of the writing device, pen tilt and to suppress extra data.

### 2.3   Core-zone detection

For difficult slanted words, slant correction techniques do not yield good results [5]. Likewise, to segment horizontally overlapped characters, we introduced a simple concept of word core-zone height. Core-zone is the area that lies between the lower baseline and upper baseline of the word. Again a very simple and fast technique is proposed to calculate lower and upper baselines. From top left point, count number of white pixels row by row. Calculate the average of those rows for which there is a change in count between current and preceding row until the first big change occurs. An average row represents the upper baseline. The same procedure is adopted for the lower baseline but from bottom to top. Results of the core zone detection for difficult words are shown in Figure 2.

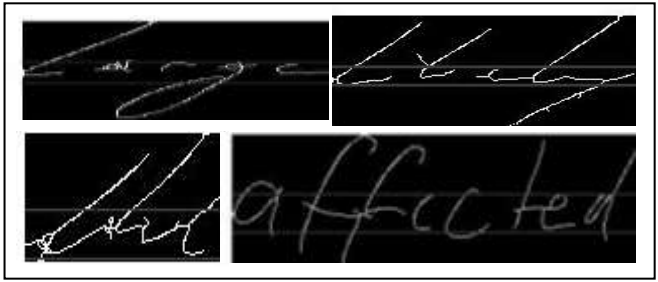

Figure 2. Core-zone height detection

### 2.4 Segmentation approach.

Proposed segmentation algorithm for difficult cursive handwritten words is presented as follow.

Step 1. Take difficult word image from IAM database.
Step 2. Perform pre-processing.
Step 3. Determine core-zone height.
Step 4. In the core zone, calculate the sum of foreground pixels (white pixels) for each column. Save those columns as candidate segment column (CSC) for which sum is 0 or 1.
Step 5. By the previous step, we have more candidate segmentation columns than actually required. A threshold is selected empirically from candidate segment columns to come out with correct segment columns. A threshold is a constant value that is derived after a number of experiments to avoid over-segmentation.

Due to the simplicity of the proposed segmentation technique, it is very fast and performs well in most of the cases. For a few characters such as w, u, v, m, n etc over segmentation occurs and this technique fails to find accurate character boundaries. However, this problem is solved heuristically by selecting a threshold and hard stroke features.

## 3  Results and analysis

### 3.1  Benchmark dataset

To exhibit an efficiency of the proposed approach, difficult slanted and horizontally overlapped words are selected from a benchmark database IAM V3.0 [58]. A few samples are shown in Figure 3.

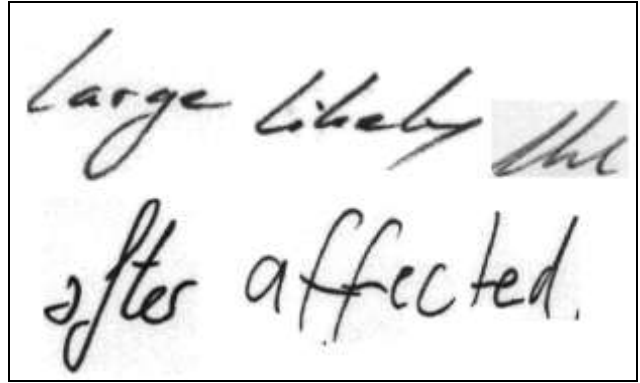
Figure 3. Difficult word samples from IAM database [58].

Two sets (317 words total) of experiments are performed. Firstly, for normal words secondly, for difficult words extracted from the benchmark database [13]. For normal words, segmentation is accurate without core-zone concept while for difficult words we faced the problem of miss-segmentation due to overlapping of characters as shown in Figure 4(a). Similarly, the problem of over-segmentation is common for the normal and difficult words without introducing threshold as depicted in Figure 4(b).

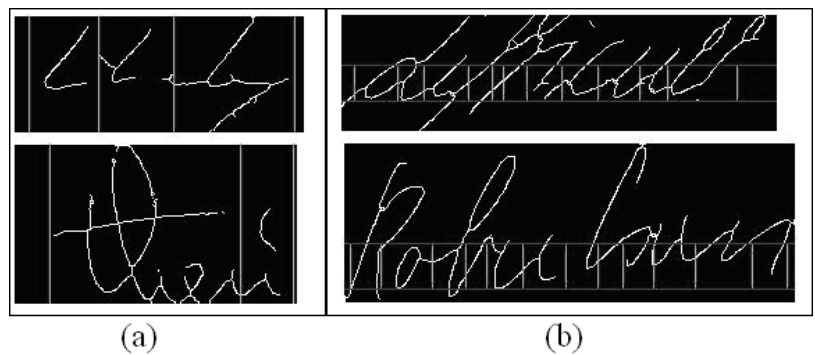
Figure 4. Miss and over-segmentation results

On analyzing the results, it is found that miss-segmentation occurred due to overlapped characters in difficult words and therefore accurate segment points were rejected as they contained more than one foreground pixels. Likewise, over-segmentation occurred due to more candidate columns containing one foreground pixels.
Few segmentation results, for normal words and for difficult words (with the core-zone concept) are presented in figure 5 and figure 6 respectively.

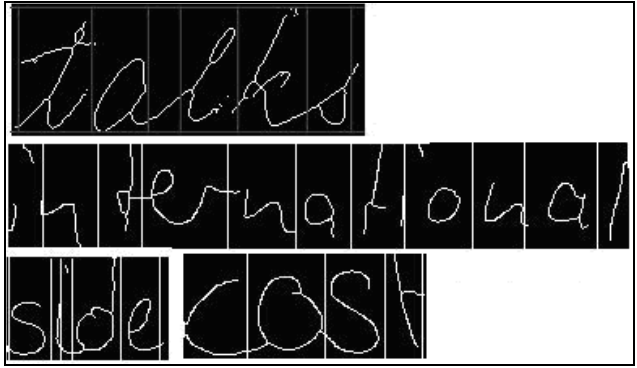

Figure 5. Segmentation results for normal words

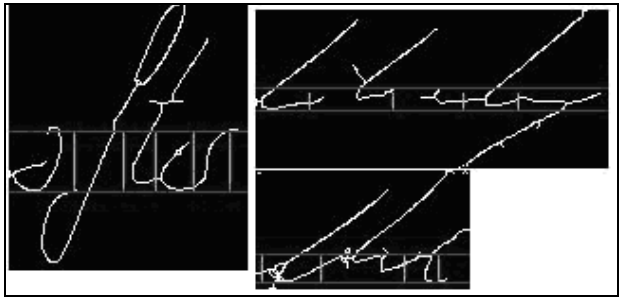

Figure 6. Segmentation results for difficult words

In order to measure the performance of the proposed approach valid, invalid and over-segmented characters are counted. There is valid detection when the algorithm finds correct character boundaries. Invalid segmentation, when the algorithm finds inaccurate character boundaries and over-segmentation when it cut a character more than desired boundaries.

Segmentation approach without introducing core zone concept performed well for normal handwritten cursive words but segmentation rate decreased significantly for the difficult words. Hence for difficult words, the core-zone concept is introduced.

For 317 (1487 characters) difficult words, 1369 characters were segmented correctly. Few characters such as u, v, w, m, n are over-segmented and miss segmentation occurs for touching/ badly overlapped characters. Segmentation results are presented in Table 1.

Table 1: Segmentation results.

| Valid segmented words | 92.06% |
|---|---|
| Miss segmentation | 1.64 % |
| Over-segmentation | 2.87 % |

**3.2 Discussion and comparison**

For the sake of time, a brief comparison of achievement for segmentation rate is presented. Verma and Gader [59] obtained 91% segmentation rate using neuro-feature based approach on words taken from CEDAR, however, the number of words was not mentioned. Likewise, Blumenstein and Verma[60] claimed 78.85% segmentation results using neuro-feature based approach for words taken from CEDAR without mentioning a number of words. In the same way, Verma [61] claimed 84.87 % accuracy for segmenting 300 CEDAR words based on neuro feature-based approach. Similarly, Chen et al [62] used neuro-feature based approach and got 95.27 % segmentation rate from 317 CEDAR words. Finally, Cheng and Blumenstein [63] using neuro-enhanced feature-based approach obtained 84.19 % segmentation rate for 317 CEDAR words. Segmentation results available in the literature are compared to the results in hand in Table 2.

## 4 Conclusion

Slant correction techniques do not yield good results for highly slanted and horizontally overlapped words. In this regard, this paper has presented a simple and effective core-zone based word segmentation approach without employing any slant correction to avoid complexity and to enhance speed. For experiments, 317 difficult words are extracted from the IAM benchmark database. Segmentation results thus obtained are promising in state of art and exhibit high speed due to its simplicity. In the future major experiment with significantly large data set shall be performed.

Table 2: Segmentation results.

| Author | Segment method | Segment rate (%) | Database |
| --- | --- | --- | --- |
| Verma and Gader [59] | Feature based + ANN | 91 | CEDAR |
| Blumenstein and Verma [60] | Feature based+ ANN | 78.85 | CEDAR |
| Verma [61] | Feature based + ANN | 84.87 | CEDAR 300 words |
| Cheng et al [62] | Feature based + ANN | 95.27 | CEDAR 317 words |
| Cheng and Blumenstein [63] | Enhanced feature based+ ANN | 84.19 | CEDAR 317 words |
| Proposed Approach | Heuristics and hard stroke features | 92.06 | IAM 317 words |